\documentclass{article}
\usepackage{ijcai23}

\usepackage{times}
\usepackage{soul}
\usepackage{url}
\usepackage[hidelinks]{hyperref}
\usepackage[utf8]{inputenc}
\usepackage[small]{caption}
\usepackage{graphicx}
\usepackage{amsmath}
\usepackage{amsthm}
\usepackage{booktabs}
\usepackage{algorithm}
\usepackage{algorithmic}
\usepackage[switch]{lineno}

\usepackage{multirow}
\usepackage{makecell}
\usepackage{array}
\usepackage{graphicx}
\usepackage{float}
\usepackage{caption}

\usepackage{subfigure}
\usepackage{arydshln}

\usepackage{CJKutf8}

\title{Clickbait Detection via Large Language Models}

\author{
Han Wang$^1$\and
Yi Zhu$^1$$^{(*)}$\and
Ye Wang$^1$\and
Yun Li$^1$\and
Yunhao Yuan$^{1}$\And
Jipeng Qiang$^1$
\affiliations
$^1$School of Information Engineering, Yangzhou University, Yangzhou, China
\emails
zhuyi@yzu.edu.cn,
wanghanhan0102@163.com,
wangye\_lj@163.com,
\{liyun, yhyuan, jpqiang\}@yzu.edu.cn
}

\begin{document}
\begin{CJK}{UTF8}{gbsn}

\maketitle

\begin{abstract}

  Clickbait, which aims to induce users with some surprising and even thrilling headlines for increasing click-through rates, permeates almost all online content publishers, such as news portals and social media. Recently, Large Language Models (LLMs) have emerged as a powerful instrument and achieved tremendous success in a series of NLP downstream tasks. However, it is not yet known whether LLMs can be served as a high-quality clickbait detection system. In this paper, we analyze the performance of LLMs in the few-shot and zero-shot scenarios on several English and Chinese benchmark datasets. Experimental results show that LLMs cannot achieve the best results compared to the state-of-the-art deep and fine-tuning PLMs methods. Different from human intuition, the experiments demonstrated that LLMs cannot make satisfied clickbait detection just by the headlines.

\end{abstract}

\section{Introduction}

With the rapid development of online applications, some content publishers try to utilize clickbait for generating profits \cite{chen2015misleading}. Clickbait refers to deliberately enticing users to click with some curious and chilling headlines, which are always unrelated to the real content or even the advertising promotion \cite{chakraborty2016stop}. The popularity of clickbait will inevitably lead to the experience degradation or even the disgust of users, there is an urgent demand to develop effective automatic clickbait detection methods \cite{liu2022clickbait}.

In the last decade, the research methods for clickbait detection evolved from feature engineering to neural networks and, more recently, into Pre-trained Language Models(PLMs). Feature engineering methods extracted features such as semantic and linguistic features for detection \cite{biyani20168,wei2017learning}. The deep neural network methods can learn more abstract and higher-level features by disentangling explanatory factors of variations behind news titles and content for clickbait detection \cite{yoon2019detecting,zheng2021deep}. Recently, fine-tuning PLMs methods, such as BERT \cite{devlin2018bert}, has shown superiority in clickbait detection tasks. However, feature engineering-based methods and deep neural networks typically require large-scale labeled data since the detection is regarded as a classification method. In the PLMs methods, the huge gap between pre-training and fine-tuning prevents detection tasks from fully utilizing pre-training knowledge.

More recently, Large Language Models (LLMs) have demonstrated the powerful ability in various NLP downstream tasks \cite{brown2020language,chowdhery2022palm,thoppilan2022lamda}, which can achieve awesome performance even in the few-shot or even zero-shot scenarios. Nevertheless, it remains unclear how LLMs perform in clickbait detection tasks compared to current methods. To address this issue, in this paper, we conduct a systematic evaluation of the few-shot and zero-shot learning capabilities of LLMs. The experiments were validated on English and Chinese open datasets, and we conducted a performance comparison between ChatGPT with the GPT-3.5 model (gpt-3.5-turbo) and GPT-4 model, and other state-of-the-art methods such as prompt-tuning.

To the best of our knowledge, this is the first attempt to validate the performance of LLMs on clickbait detection, we provide a preliminary evaluation including detection results and robustness. The key findings and insights are summarized as follows:

\begin{itemize}
	\item LLMs achieves unsatisfying results in the few-shot and zero-shot scenarios compared to the state-of-the-art clickbait detection methods. We found that fine-tuning PLMs and prompt-tuning can achieve better results just based on the titles, which is consistent with humans in the real-world.
	\item ChatGPT is a monolithic model capable of supporting multiple languages, which makes it a comprehensive multilingual clickbait detection technique. After evaluating the performance of ChatGPT on the task of clickbait detection across two languages (English and Chinese), we observed that it achieved stable results on almost all evaluation metrics. This also confirms that LLMs can be adapted to other languages.
	%\item After evaluating the results across two languages (English and Chinese), we observed that the prompt-tuning methods can significantly and consistently outperform ChatGPT, we believe that prompt-tuning can leverage chain-of-thought for achieving better performance on clickbait detection tasks.
\end{itemize}

It is worth mentioning that the source code and all the results of this paper are available at https://github.com/zhuyiYZU/chatGPTforClickbait.

\section{Related Work}

\subsection{Clickbait Detection}

Clickbait detection is an emerging field that has attracted increasing attention in recent years. On most online services, such as e-commerce, social media, and news portals, more clicks mean more profit and commercial revenue. Early clickbait detection methods mainly focus on extracting a variety of features for detection tasks, such as semantic \cite{rony2017diving}, linguistics \cite{blom2015click}, and multi-modal features \cite{chen2015misleading}. However, these methods require expert knowledge for feature selection, and the handcrafted features are limited in representing more abstract and higher-level information.

In recent years, deep neural network models facilitate crossing and combination among even more diverse and sophisticated features, which has shown fairly good performance in clickbait detection. The popular deep neural networks, such as Recurrent Neural Networks (RNN) \cite{anand2017we}, Convolutional Neural Networks (CNN) \cite{agrawal2016clickbait}, Attention Mechanism \cite{mishra2020musem}, and Graph Attention Networks \cite{liu2022clickbait}, have already been devoted to the clickbait detection tasks. Despite the success of deep learning methods, due to the requirements on large-scale labeled training datasets, lead to high costs in collecting eligible training data. Recently, some domain adaptation \cite{lopez2018hybridizing} and data augmentation \cite{yoon2019detecting} methods have been proposed to address the issues, however, these methods may bring additional noise in detection tasks.

More recently, pre-trained language models (PLMs) such as BERT \cite{devlin2018bert}, RoBERTa \cite{liu2019roberta}, and T5 \cite{raffel2020exploring} have emerged as a powerful instruments for language understanding and generation. Through the fine-tuning PLMs on the special downstream task, the rich knowledge distributed in PLMs can be stimulated to better serve downstream tasks including clickbait detection \cite{indurthi2020predicting,yi2022clickbait}. Despite the success of fine-tuning PLMs, some recent studies find one of its critical challenges is the significant gap of objective forms in pre-training and fine-tuning, which restricts taking full advantage of knowledge in PLMs.	

\subsection{Large Language Models}

Represented by GPT-4 \cite{brown2020language}, Large Language Models (LLMs) have achieved superior performance, especially in few-shot learning scenarios \cite{brown2020language,chowdhery2022palm,thoppilan2022lamda}. Different from the previous PLMs methods, LLMs has two distinct advantages. The first is the larger scale, LLMs have a much larger scale in terms of model parameters and training data. Secondly, without fine-tuning PLMs, LLMs can prompt few-shot learning that requires no additional neural layer and shows excellent performance. However, there is no work on the capabilities of LLMs on clickbait detection tasks.

\section{Methodology}

Through the rich knowledge in upstream large models, they can achieve excellent performance in downstream tasks with low resources. Considering the lack of large-scale training corpus for clickbait detection tasks, we will test the performance of LLMs in few-shot and zero-shot scenarios on clickbait detection.

LLMs typically use prompts (i.e. specific templates) to guide the model in predicting output or answers, without requiring specific training on the data. Utilizing this form of prompt, we conducted experiments with different prompts on OpenAI's most representative available models GPT3.5 (gpt-3.5-turbo) and GPT-4 for ChatGPT.

\subsection{Prompt}To stimulate the rich knowledge distributed in LLMs, we manually designed the prompts for validating the performance of clickbait detection. The details of the prompts are illustrated in Table 1. For the two prompts, \{Clickbait Sentence\} and \{Not-Clickbait Sentence\} refer to the input sentence, \{Yes, it is a clickbait\} and \{No, it is not a clickbait\} are the corresponding labels respectively. \{Results\} is that place that carries the outputs of LLMs.

\begin{table}[htbp]
	\caption{The hand-crafted prompts for clickbait detection.}
	\centering
	\setlength{\tabcolsep}{3pt}
	\begin{tabular}{ll}
		\toprule
		P\_id &Prompts\\
		%\multicolumn{2}{c}{Prompts}\\
		\midrule
		\multirow{8}{*}{P1}  &  I want you to detect whether the input sentence is clickbait or not.\\
		~   &   Input: \textbf{\{Clickbait Sentence\}} \\
		~   &   Output: \textbf{\{Yes, it is a clickbait\}} \\
		~   &   Input: \textbf{\{Not-Clickbait Sentence\}} \\
		~   &   Output: \textbf{\{No, it is not a clickbait\}} \\
		~   &   ... \\
		~   &   Input: \textbf{\{Clickbait or Not-Clickbait Sentence\}} \\
		~   &   Output: \textbf{\{Results\}} \\
		\midrule
		\multirow{10}{*}{P2} & Sentence:\textbf{\{Clickbait Sentence\}} \\
		~   &   Question: Detect the above sentence is clickbait or not.\\
		~   &   Answer: \textbf{\{Yes, it is a clickbait\}} \\
		~   &   Sentence:\textbf{\{Not-Clickbait Sentence\}} \\
		~   &   Question: Detect the above sentence is clickbait or not.\\
		~   &   Answer: \textbf{\{No, it is not a clickbait\}} \\
		~   &   ... \\
		~   &   Sentence:\textbf{\{Clickbait or Not-Clickbait Sentence\}} \\
		~   &   Question: Detect the above sentence is clickbait or not.\\
		~   &   Answer: \textbf{\{Results\}} \\
		\bottomrule
	\end{tabular}
\end{table}

In the first prompt (P1), the \{guide-input-output\} pattern is employed to guide LLMs for detecting clickbaits. In the second prompt (P2), the \{sentence-question-answer\} pattern is utilized to detect clickbait in the form of a question. It is worth mentioning that the specialized guide words, such as "Output:" and "Answer:", are added to the prompts, then the clickbait detection is regarded as the close-style task for LLMs, which ensures to achievement of a unique output. When performing multilingual clickbait detection tasks, we translate these two prompts into Chinese used in specific tasks .

\section{Experiments}	
\subsection{Datasets}
To evaluate the performance of LLM for clickbait detection in English and Chinese, we conducted experiments on seven datasets. The DL-Clickbait (DLC), Clickbait news detection (CND), and SC-Clickbait (SCC) are three well-known public English datasets. Sina, Tencent, WeChat, and Paper are four Chinese clickbait detection datasets. The statistical details of all the datasets are presented in Table 2.

\begin{table*}[htbp]
	\caption{Statistics of the news datasets. "\#" and "avg.\#" denote "the number of" and "the average number of".}
	\centering
	\resizebox{0.7\linewidth}{!}{
		\begin{tabular}{cccccccc}
			\toprule
			Statistic&DLC&CND&SCC&Sina&Tencent&WeChat&Paper\\
			\midrule
			\#total headline&2388&8324&32000&1648&2594&2808&30800\\
			\#train sets&1671&5801&22400&1153&1815&1965&21560\\
			\#test sets&717&2523&9600&495&779&843&9240\\
			avg.\# words per sentence&67&13&53&604&316&195&621\\
			\bottomrule
		\end{tabular}
	}
	%\label{tab:booktabs}
\end{table*}

\subsection{Baseline}
We compare our ChatGPT with the following deep neural network and fine-tuning PLMs methods.

\textbf{CNN} \cite{agrawal2016clickbait} CNN utilizes kernels with different sizes to find various features from the input text. These features are automatically detected and are used to train a neural network for the corresponding task. It understands the input text from different perspectives, and can be applied to various nature language processing tasks.

\textbf{BiLSTM} \cite{kumar2018identifying} A bidirectional LSTM with an attention mechanism to learn the extent to which a word contributes to the clickbait score of a social media post in a different way. It used a Siamese net to detect similarities between the source and target data.

\textbf{GATED\_CNN} \cite{yang2019fake} The method can be stacked to represent large context sizes and extract hierarchical features over larger and larger contexts with more abstractive features.

\textbf{FAST\_TEXT} \cite{joulin2016bag} FAST\_TEXT is a text-classification method, and the sequence of words is considered in it, which is a model learning distributed representations of words based on ordered words.

\textbf{BERT} \cite{devlin2018bert} The BERT method uses words and sentences to distinguish the context of words in a sentence.

\textbf{MFWCD-BERT} \cite{liu2022clickbait} The multi-feature WeChat clickbait detection method, which incorporates semantic, syntactic, and auxiliary information, achieves excellent performance. It leverages an enhanced Graph Attention Network (GAT) to aggregate title and local syntactic structures, using an attention mechanism to capture valuable structural features.

\textbf{PT} \cite{liu2023pre} PT is a technique known as prompt tuning, which incorporates a cloze-style task during the model fine-tuning process. This approach enables a more effective utilization of pre-training information, thereby further enhancing the model's performance in few-shot learning scenarios.

\textbf{PEPT} \cite{wu2023detecting} PEPT is a method that introduces prompt learning into clickbait detection. It effectively performs classification tasks and part-of-speech tagging using pre-trained language models by training them on specific tasks. It's worth noting that this approach is inspired by typical grammar and semantic features of clickbait, and it achieves grammar-guided semantic understanding, which can perform well in both low-resource and full-scale scenarios.

\subsection{Implementation Details and Evaluation Metrics}

\subsubsection{Few-shot settings}

In the experiments, we randomly selected K (5,10,20) instances as the training set for our ChatGPT, and selected part of the samples as the test set. Considering that too few training samples may greatly affect the effectiveness of the baselines, we select different training samples from different datasets, where the number of training samples is 300/600/1200, 400/800/1600, and 1000/2000/4000 for English datasets (DL-Clickbait, CND, and SC-Clickbait datasets) and 200/800/1600, 400/800/1600, 400/800/1600 and 800/1600/3200 for Chinese datasets(Sina, Tencent, WeChat and Paper), corresponding to 5/10/20 shots. Considering that different choices of few-shot training during training affect the test results, we repeatedly sample the same data on K random seeds simultaneously and calculate their mean values as reported. Considering more intuitive approaches to training the ChatGPT model, here are a few specific examples. As shown in Figure 1, examples are provided for training the model using few-shot (5-shot) methods, along with the corresponding results.

\begin{figure*}[htbp]
	%\addtocunter{figure}{-1}
	\centering
	\subfigure[English P1]{
		\begin{minipage}[b]{0.45\textwidth}
			\includegraphics[scale = 0.6]{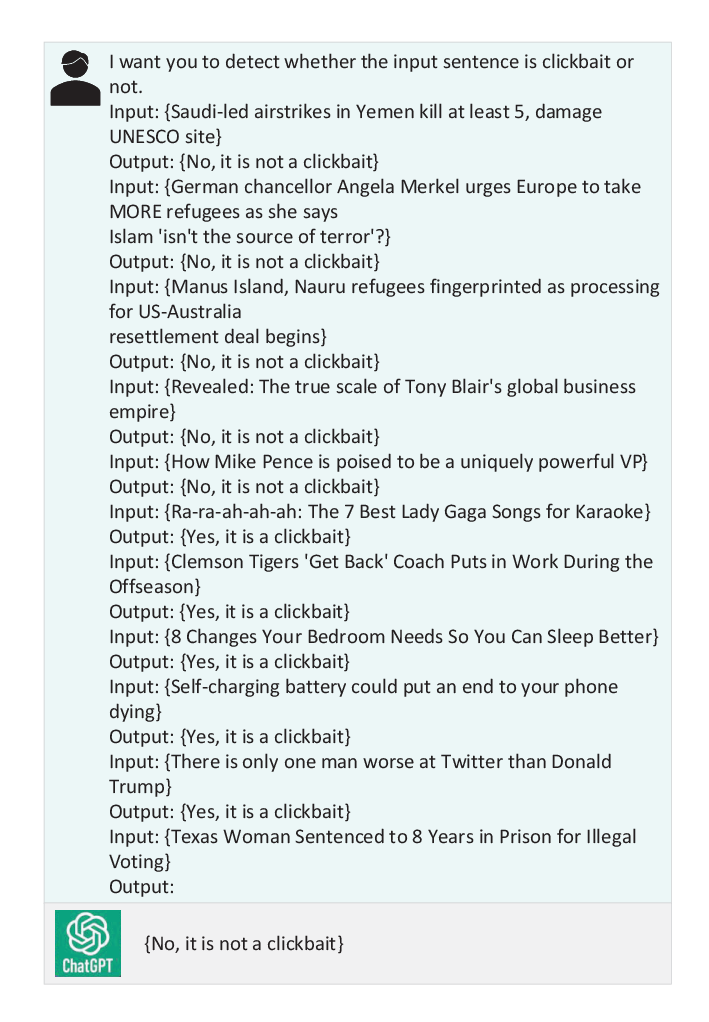}
	\end{minipage}}
	\subfigure[Chinese P1]{
		\begin{minipage}[b]{0.45\textwidth}
			\includegraphics[scale = 0.6]{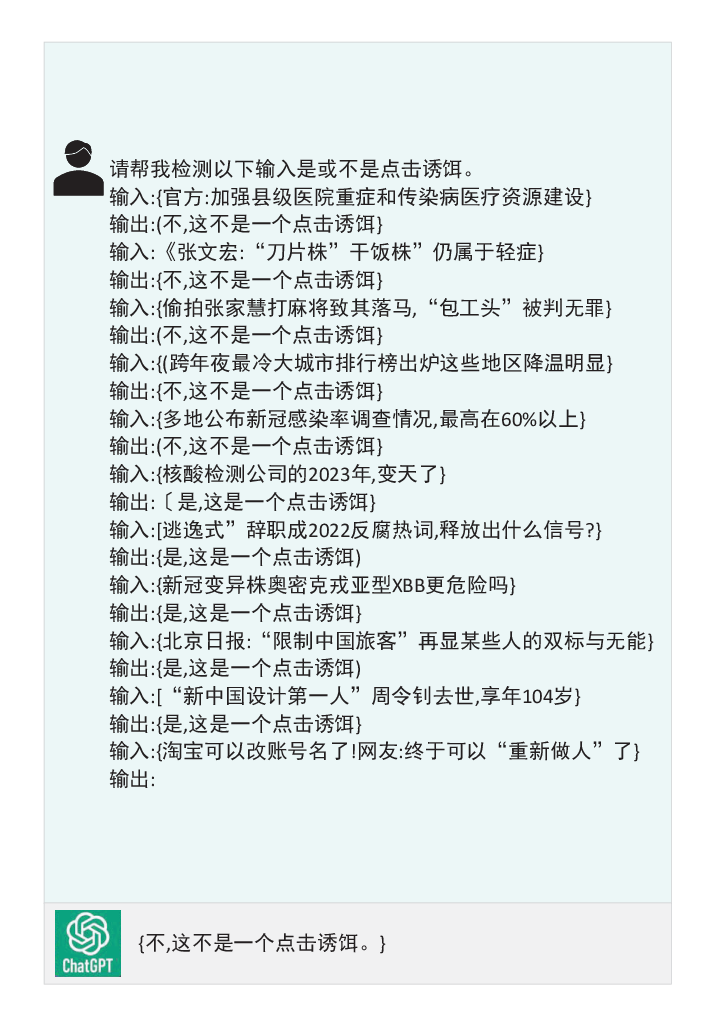}
	\end{minipage}}
	\subfigure[English P2]{
		\begin{minipage}[b]{0.45\textwidth}
			\includegraphics[scale = 0.6]{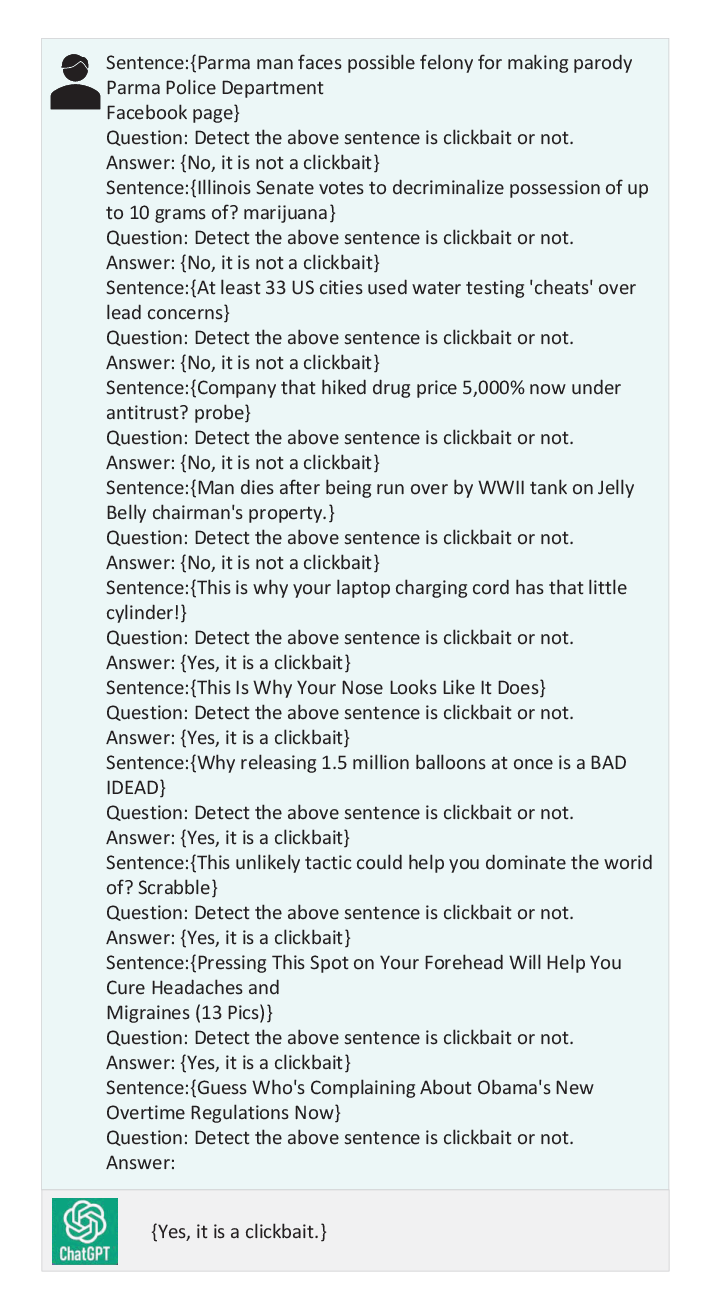}
	\end{minipage}}
	\subfigure[Chinese P2]{
		\begin{minipage}[b]{0.45\textwidth}
			\includegraphics[scale = 0.6]{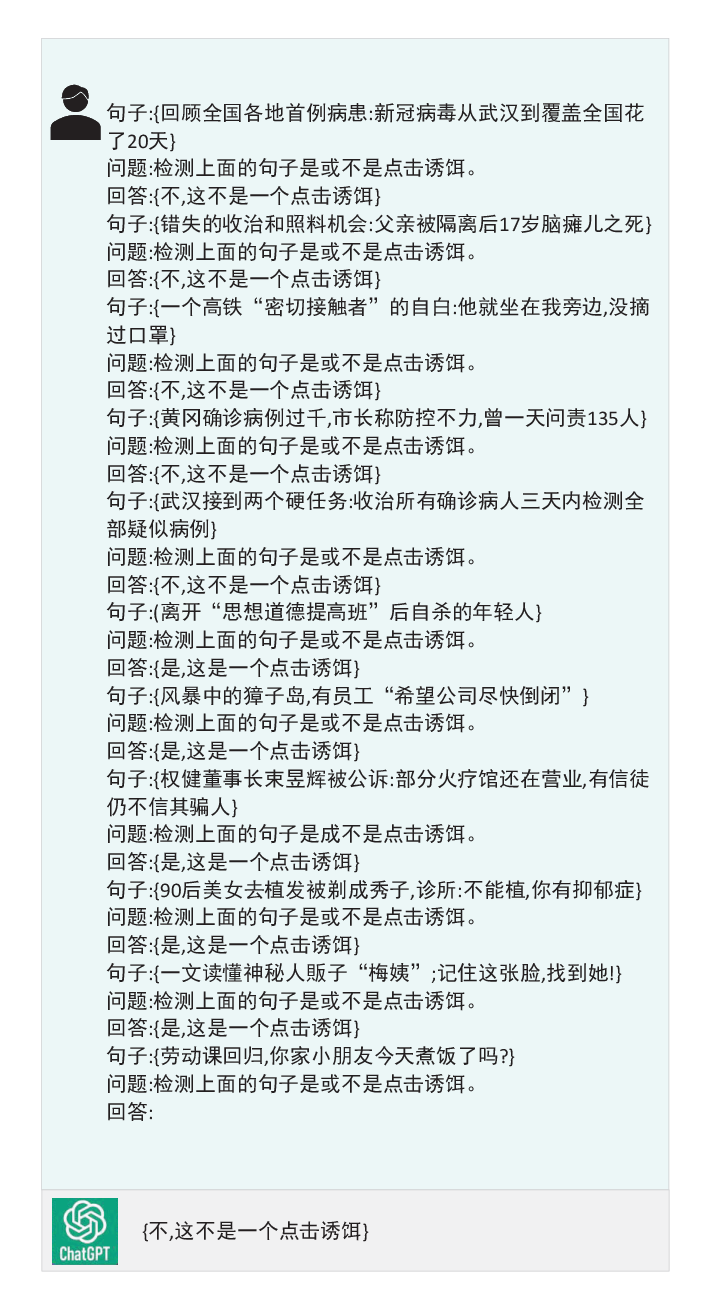}
	\end{minipage}}
	\caption{The example of GPT-3.5 detecting clickbait errors.}
	\centering
\end{figure*}

\subsubsection{Parameter settings}

%We chose the latest available LLMs GPT3.5 (gpt-3.5-turbo). In order to interact with the model, we access it by making API calls to the ChatGPT API, in which calling GPT3.5 requires payment. This model's maximum context length is 4096 tokens, approximating 100 news headlines. Moreover, this model can only call three news headlines a minute. In addition, for the few-shot experiments, we select a few sentences that ChatGPT got wrong from the seven data sets mentioned above as examples of clickbait detection. Figure 1 shows ChatGPT detecting clickbait errors.
For using ChatGPT for clickbait detection, we have chosen the most representative available models GPT3.5 (gpt-3.5-turbo) and GPT-4, to interact with. To establish communication with the model, we have opted to use the ChatGPT API. First, we need to register and apply for a ChatGPT API key from OpenAI. This API key will be used to access the API. Then, we send an HTTP POST request to the API endpoint to engage in a conversation with the ChatGPT model.In this process, we set up an array of messages, including user messages and system messages, corresponding to the two roles, "user" and "assistant". Based on the input of users, messages will be alternately added to the array as user messages and system messages. Subsequently, we extract the assistant's responses from the JSON response returned by the API.It's important to note that using the ChatGPT API incurs a cost, and the text content of API requests must not exceed the maximum token limit of the model, which, in this case, is 4096 tokens, roughly equivalent to 70 news headlines. Additionally, there are rate limits for API requests, with a maximum of three news headlines per minute and no more than two hundred requests in a day.In addition, for the few-shot experiments, we select a few sentences that GPT-3.5 got wrong from the seven data sets mentioned above as examples of clickbait detection. Figure 1 shows ChatGPT detecting clickbait errors.

We also provide a detailed description of the parameter settings for our text classification models (CNN, BiLSTM, and FastText) used in the clickbait detection task. For the CNN model, we employ three convolutional layers, each with a kernel size of 3x3 and 64, 128, and 256 kernels, respectively. The stride is set to 1, and padding is "same." Following each convolutional layer, we incorporate max-pooling layers with a kernel size of 2x2 and a stride of 2. After the convolutional layers, we have two fully connected layers with 512 and 256 neurons, respectively. Additionally, we apply ReLU activation functions after each convolutional and fully connected layer. For the BiLSTM model, we utilize two layers of bidirectional LSTM, with 64 and 128 LSTM units in each layer. Since the BiLSTM layers are bidirectional, the total number of LSTM units is 256 (128 * 2). After the BiLSTM layers, we add two fully connected layers with 512 and 256 neurons, respectively. Similarly, ReLU activation functions are applied after each fully connected layer. In the case of the FastText model, we represent the text data as bag-of-words vectors. We set up a fully connected layer with two neurons, corresponding to the binary classification task. After the fully connected layer, we use the Sigmoid activation function. For all three models, we employ binary cross-entropy loss functions as the objective functions to measure model performance. We use the Adam optimizer, with a typical training batch size of 64 and 50 training epochs. We conducted hyperparameter tuning, including learning rates, the number of convolutional kernels, LSTM unit numbers, bag-of-words sizes, and word vector dimensions, selecting the best hyperparameter values through random search.

For the BERT model, we employed the BERT-base-uncased variant and conducted training over 10 epochs using a batch size of 32. For training prompt-based learning models (PT, PEPT), we utilized the xlm-roberta-large pretrained language model. To prevent overfitting, we applied a dropout rate of 0.5 during the training process. We configured the learning rate at 3e-5 and set the batch size to 16, incorporating a weight decay of 1e-5. Additionally, we performed validation steps to fine-tune the model's hyperparameters. Across all our small-scale experiments, we maintained a consistent number of 10 epochs for comprehensive training and reliable results. We employed the Adam optimizer to optimize the model parameters.

The experimental outcomes were derived from a server configuration featuring an NVIDIA GeForce RTX 3090 Founders Edition GPU, an Intel(R) Core(TM) i9-10980XE CPU clocked at 3.00 GHz, and 125 GB of RAM.

\subsubsection{Evaluation metrics}

To test the effect of detection, we conduct four evaluation metrics to evaluate our method in experiments, such as accuracy, precision, recall and F1-score.

\textbf{accuracy}  The accuracy can be defined as the ratio of correctly predicted samples to the total number of samples.

\begin{equation}
	Acc=\frac{TP+TN}{TP+TN+FP+FN}
\end{equation}

\textbf{precision}  The positive prediction rate can be defined as the ratio of correctly predicted positive samples to the total number of positive samples in the prediction.

\begin{equation}
	Pre=\frac{TP}{TP+FP}
\end{equation}

\textbf{recall}  The positive precision can be defined as the ratio of correctly predicted positive samples to the total number of samples labeled positive.

\begin{equation}
	Rec=\frac{TP}{TP+FN}
\end{equation}

\textbf{F1-score}  The F1-score can be defined as the harmonic average of precision and recall.

\begin{equation}
	F1=\frac{{2}\times{Pre}\times{Rec}}{Pre+Rec}
\end{equation}

\begin{table*}[htbp]
	\caption{The Accuracy, Precision, Recall, F1-scores results (\%) on English datasets ( DLC, CND, and SCC). 200/5,300/10, and 600/20 denote the training number of not-prompt-tuning(CNN, BILSTM, GATED-CNN, FAST-TEXT, BERT, and MFWCD-BERT) and prompt-tuning(PT, PEPL, GPT-3.5 and GPT-4) methods. Ditto above. The bolder ones mean better.}
	\centering
	\resizebox{1.1\linewidth}{!}{
		\begin{tabular}{clcccc|cccc|cccc}
			\toprule
			Data & Method  & \multicolumn{12}{c}{Shot}  \\
			\cline{3-14}

			\multicolumn{2}{c}{ }  & \multicolumn{4}{c}{300/5}  & \multicolumn{4}{c}{600/10} & \multicolumn{4}{c}{1200/20}  \\
			\midrule
			\multicolumn{2}{c}{ } & Acc     &Pre       &Rec        & F1            & Acc     &Pre       &Rec        & F1            & Acc     &Pre       &Rec        & F1  \\
			\multirow{10}*{DLC}&CNN       & 60.39 & 70.95 & 60.39 & 65.24 & 70.29 & 68.71 & 70.29 & 69.49 & 81.72 & 81.55 & 81.72 & 81.63 \\
			&BiLSTM    & 73.08 & 72.88 & 73.08 & 72.97 & 77.68 & 77.87 & 77.68 & 77.77 & 82.28 & 82.89 & 82.28 & 82.58 \\
			&GATED-CNN & 77.80 & 80.05 & 75.41 & 78.01 & 79.34 & 79.86 & 78.16 & 78.43 & 78.39 & 79.71 & 79.93 & 79.66 \\
			&FAST-TEXT & 65.23 & 72.31 & 63.84 & 72.32 & 74.89 & 71.07 & 74.30 & 65.58 & 75.13 & 78.33 & 73.60 & 77.63 \\
			&BERT      & 71.86 & 71.83 & 71.86 & 71.84 & 77.40 & 80.62 & 77.40 & 78.97 & 82.42 & 82.13 & 82.42 & 82.27\\
			&MFWCD-BERT           &79.13   &80.44   &79.13   &78.91         &82.19   &82.21   &82.19   &82.19       &85.00   &85.90   &85.00   &84.91\\
			&PT                   &81.86   &82.69   &81.86   &80.60         &82.70   &82.43   &82.70   &82.42       &84.79   &84.60   &84.79   &84.52\\
			&PEPL                 &83.31   &82.70   &79.29   &80.53         &87.41   &90.11   &82.40   &\textbf{84.74}       &\textbf{88.97}   &\textbf{89.58}   &\textbf{85.67}   &\textbf{87.16}\\
			\cdashline{2-14}
			&GPT-3.5              &80.75   &69.06   &\textbf{82.79}   &73.56         &79.33   &69.51   &\textbf{88.02}   &70.88       &82.07   &75.66   &82.64   &73.67\\
			&GPT-4   & \textbf{90.54} & \textbf{94.00}    & 77.05 & \textbf{84.68} & \textbf{89.96} & \textbf{96.24} & 73.36 & 83.26 & 86.05 & 88.3  & 68.03 & 76.85 \\
			\midrule
			
			\multicolumn{2}{c}{ }  & \multicolumn{4}{c}{300/5}  & \multicolumn{4}{c}{600/10} & \multicolumn{4}{c}{1200/20}  \\
			\midrule
			\multicolumn{2}{c}{ } & Acc     &Pre       &Rec        & F1            & Acc     &Pre       &Rec        & F1            & Acc     &Pre       &Rec        & F1  \\
			\multirow{10}*{CND}&CNN       & 73.74          & 69.70          & 71.57 & 68.82 & 76.47 & 74.83          & 74.33 & 75.70          & 73.19          & 72.05          & 75.27          & 66.61          \\
			&BiLSTM    & \textbf{75.18} & \textbf{75.78} & 74.06 & 73.50 & 75.74 & 79.54          & 78.28 & \textbf{79.15} & 71.79          & 81.70          & 80.08          & 82.71          \\
			&GATED-CNN & 70.92          & 71.41          & 70.18 & 67.69 & 71.26 & 74.53          & 76.04 & 72.28          & 68.00          & 74.54          & 74.23          & 75.39          \\
			&FAST-TEXT & 54.18          & 66.01          & 64.12 & 61.22 & 72.51 & 71.61          & 69.57 & 53.51          & 79.25          & 77.76          & 74.47          & 79.69          \\
			&BERT      & 73.17          & 73.19          & 73.17 & 73.16 & 76.90 & \textbf{80.65} & 76.90 & 76.71          & \textbf{83.28} & \textbf{85.69} & \textbf{83.02} & \textbf{83.80}\\
			&MFWCD-BERT           &75.07   &75.18   &\textbf{74.98}   &\textbf{74.99}         &\textbf{78.09}   &79.54   &\textbf{79.11}   &78.97       &80.58   &81.89 &80.16   &80.94\\
			&PT                   &53.07   &54.04   &53.07   &50.05         &62.89   &63.34   &62.89   &62.57       &66.58   &68.09  &66.58   &65.86\\
			&PEPL                 &58.19   &61.50   &58.19   &54.95         &62.73   &64.47   &62.73   &61.58       &63.08   &63.46  &63.08   &62.83\\
			\cdashline{2-14}
			&GPT-3.5              &60.00   &59.02   &68.36   &62.06         &59.44   &60.76   &61.00   &60.13       &60.90   &63.33   &66.88   &62.57\\
			&GPT-4  & 68.61 & 69.47 & 66.97 & 68.2  & 64.32 & 64.65 & 63.2  & 63.92 & 63.47 & 72.56 & 43.32 & 54.25 \\
			\midrule
			
			\multicolumn{2}{c}{ }  & \multicolumn{4}{c}{1000/5}  & \multicolumn{4}{c}{2000/10} & \multicolumn{4}{c}{4000/20}  \\
			\midrule
			\multicolumn{2}{c}{ } & Acc     &Pre       &Rec        & F1            & Acc     &Pre       &Rec        & F1            & Acc     &Pre       &Rec        & F1  \\
			\multirow{10}*{SCC}&CNN       & 75.31 & 76.30 & 75.31          & 75.80 & 82.28 & 82.01 & 82.28 & 82.14 & 86.19 & 86.11 & 86.19 & 86.14 \\
			&BiLSTM    & 79.77 & 79.36 & 79.77          & 79.55 & 83.22 & 83.54 & 83.22 & 83.37 & 87.09 & 87.86 & 87.09 & 87.47 \\
			&GATED-CNN & 92.07 & 91.78 & \textbf{91.90} & 91.67 & 93.56 & 93.14 & 93.27 & 93.56 & 93.22 & 94.67 & 94.28 & 94.45 \\
			&FAST-TEXT & 84.09 & 83.66 & 83.18          & 82.52 & 85.99 & 85.76 & 85.27 & 85.74 & 86.67 & 86.96 & 86.60 & 86.68 \\
			&BERT      & 78.52 & 78.04 & 78.52          & 78.27 & 83.96 & 83.78 & 83.96 & 83.86 & 89.92 & 89.95 & 89.92 & 89.93\\
			&MFWCD-BERT           &81.94   &81.95   &81.94   &81.94         &83.82   &83.91   &83.82   &83.81       &85.69   &86.10   &85.69   &85.65\\
			&PT                   &90.91   &91.12   &90.91   &90.90         &92.94   &92.45   &92.47   &92.42       &93.37   &93.44   &93.37   &93.37\\
			&PEPL                 &83.10   &85.57   &83.11   &82.81         &95.00   &95.00   &95.00   &95.00        &96.53   &96.54   &\textbf{96.53}   &96.53\\
			\cdashline{2-14}
			&GPT-3.5              &83.94   &88.63   &90.01   &83.08         &86.14   &92.41   &86.93   &85.09       &88.73   &95.54   &87.81   &87.82\\
			&GPT-4  & \textbf{94.62} & \textbf{98.83} & 90.3  & \textbf{94.37} & \textbf{96.88} & \textbf{97.92} & \textbf{95.80}  & \textbf{96.85} & \textbf{97.17} & \textbf{98.79} & 95.5  & \textbf{97.12}\\
			\bottomrule
		\end{tabular}
	}
	
	%\label{tab:booktabs}
\end{table*}

\begin{table*}[htbp]
	\caption{The Accuracy, Precision, Recall, F1-scores results (\%) on Chinese datasets (Sina, Tencent, WeChat, and Paper). 200/5,400/10, and 800/20 denote the training number of not-prompt-tuning(CNN, BILSTM, GATED-CNN, FAST-TEXT, BERT, and MFWCD-BERT) and prompt-tuning(PT, PEPL, GPT-3.5 and GPT-4) methods. Ditto above. The bolder ones mean better.}
	\centering
	\resizebox{1.1\linewidth}{!}{
		\begin{tabular}{clcccc|cccc|cccc}
			\toprule
			Data & Method  & \multicolumn{12}{c}{Shot}  \\
			\cline{3-14}
			\multicolumn{2}{c}{ }  & \multicolumn{4}{c}{200/5}  & \multicolumn{4}{c}{800/10} & \multicolumn{4}{c}{1600/20}  \\
			\midrule
			\multicolumn{2}{c}{ } & Acc     &Pre       &Rec        & F1            & Acc     &Pre       &Rec        & F1            & Acc     &Pre       &Rec        & F1  \\
			\multirow{10}*{Sina}&CNN       & 64.52 & 59.23 & 64.52 & 61.76 & 69.62 & 67.09 & 69.62 & 68.33 & 79.51 & 77.40 & 79.51 & 78.44 \\
			&BiLSTM    & 70.04 & 70.26 & 70.04 & 70.15 & 74.48 & 74.38 & 74.48 & 74.43 & 81.72 & 80.98 & 81.72 & 81.35 \\
			&GATED-CNN & 73.87 & 72.09 & 71.25 & 71.37 & 75.61 & 76.07 & 77.35 & 75.66 & 77.00 & 76.20 & 75.61 & 76.93 \\
			&FAST-TEXT & 56.10 & 62.23 & 52.79 & 63.94 & 74.04 & 71.45 & 72.13 & 72.74 & 73.17 & 73.76 & 74.04 & 71.81 \\
			&BERT      & 52.36 & 52.86 & 44.25 & 48.20 & 55.25 & 57.69 & 46.51 & 51.50 & 65.96 & 71.42 & 55.56 & 62.50\\
			&MFWCD-BERT           &77.80   &79.49   &77.41   &77.24         &82.43   &85.15   &\textbf{82.93}   &\textbf{82.39}       &85.28   &\textbf{88.56}  &85.22  &85.77\\
			&PT                   &\textbf{79.14}   &79.17  &\textbf{79.13}  &\textbf{78.91}         &\textbf{82.54}   &82.65  &82.58   &82.11       &\textbf{86.05}   &86.12  &\textbf{86.17}   &\textbf{86.08}\\
			&PEPL                 &77.51   &80.04  &77.24   &77.34         &79.28   &79.17  &79.13   &79.20       &84.92   &85.44  &84.62   &84.01\\
			\cdashline{2-14}
			&GPT-3.5              &70.73   &76.21   &60.28   & 67.32         &69.16   &71.65   &63.41   &67.28       &75.30   &83.80   & 62.85 &71.83\\
			&GPT-4  & 73.87 & \textbf{93.08} & 51.57 & 66.37 & 78.36 & \textbf{91.33} & 62.59 & 74.27 & 75.61 & 82.67 & 64.81 & 72.66\\
			\midrule
			
			\multicolumn{2}{c}{ }  & \multicolumn{4}{c}{400/5}  & \multicolumn{4}{c}{800/10} & \multicolumn{4}{c}{1600/20}  \\
			\midrule
			\multicolumn{2}{c}{ } & Acc     &Pre       &Rec        & F1            & Acc     &Pre       &Rec        & F1            & Acc     &Pre       &Rec        & F1  \\
			\multirow{10}*{Tencent}&CNN       & 68.50 & 69.31 & 68.50 & 68.91 & 74.43 & 73.78 & 74.43 & 74.11 & 80.19 & 81.08 & 80.19 & 80.63 \\
			&BiLSTM    & 71.49 & 73.48 & 71.49 & 72.47 & 74.51 & 73.90 & 74.51 & 74.22 & 82.30 & 82.60 & 82.30 & 82.46 \\
			&GATED-CNN & 55.37 & 56.28 & 56.54 & 55.63 & 60.60 & 60.29 & 57.33 & 55.71 & 60.08 & 60.74 & 57.98 & 60.31 \\
			&FAST-TEXT & 53.53 & 50.58 & 51.05 & 56.49 & 56.28 & 57.75 & 57.07 & 49.95 & 55.37 & 57.94 & 58.12 & 58.20 \\
			&BERT      & 79.28 & 78.93 & 79.28 & 79.11 & 82.66 & 83.40 & 82.66 & 83.03 & 85.52 & 83.22 & 85.52 & 84.38\\
			&MFWCD-BERT           &\textbf{81.44}   &\textbf{86.34}   &\textbf{81.40}   &\textbf{81.02}         &\textbf{83.78}   &84.01   &\textbf{83.70}   &\textbf{83.41}       &\textbf{87.71}   &\textbf{88.03}  &\textbf{87.61}  &\textbf{87.20}\\
			&PT                   &79.72   &82.48   &79.82   &79.46         &81.76   &\textbf{84.48}   &81.35 &78.62       &82.59   &82.65  &82.58  &82.54\\
			&PEPL                 &74.87   &74.90   &74.82   &74.70         &76.72   &77.03   &76.65   &76.61       &80.03   &80.05   &80.00   &80.81\\
			\cdashline{2-14}
			&GPT-3.5              &59.34   &57.91   &71.13   &63.84          &56.02   &54.71   &69.90   &61.38       &59.08   &57.17   &71.99 &63.73\\
			&GPT-4   & 55.97 & 67.76 & 25.94 & 37.52 & 58.32 & 66.16 & 34.29 & 45.17 & 60.99 & 60.14 & 65.18 & 62.56\\
			\midrule
			
			\multicolumn{2}{c}{ }  & \multicolumn{4}{c}{400/5}  & \multicolumn{4}{c}{800/10} & \multicolumn{4}{c}{1600/20}  \\
			\midrule
			\multicolumn{2}{c}{ } & Acc     &Pre       &Rec        & F1            & Acc     &Pre       &Rec        & F1            & Acc     &Pre       &Rec        & F1  \\
			\multirow{10}*{WeChat}&CNN       & 77.08 & 78.70 & 77.08 & 77.88 & 82.05 & 80.34 & 82.05 & 81.19 & 83.35 & 81.02 & 83.35 & 82.17 \\
			&BiLSTM    & 78.86 & 77.41 & 78.86 & 78.13 & 83.80 & 83.53 & 83.80 & 83.67 & 85.20 & 85.78 & 85.20 & 85.49 \\
			&GATED-CNN & 67.74 & 69.17 & 69.17 & 67.62 & 83.45 & 86.31 & 84.29 & 85.27 & 87.74 & 88.16 & 86.67 & 86.02 \\
			&FAST-TEXT & 58.10 & 62.59 & 60.48 & 61.56 & 81.07 & 83.26 & 81.79 & 83.13 & 84.17 & 84.60 & 84.29 & 81.18 \\
			&BERT      & 83.19 & 79.38 & 83.19 & 81.24 & 85.11 & 84.95 & 85.11 & 85.03 & 88.14 & 89.00 & 88.14 & 88.57\\
			&MFWCD-BERT           &\textbf{85.92}   &\textbf{89.12}   &85.81   &\textbf{85.76}         &\textbf{88.86}   &\textbf{89.59}   &88.79  &\textbf{88.31}       &\textbf{90.31}   &\textbf{92.12}  &90.58  &\textbf{91.21}\\
			&PT                   &78.27   &78.54   &78.27   &78.25         &81.37   &82.34   &81.40   &81.32       &83.96   &84.72   &83.96  &83.87\\
			&PEPL                 &76.35   &81.10   &76.37    &76.18         &80.11   &82.27   &80.45   &80.25       &84.55   &86.61   &84.26   &84.03\\
			\cdashline{2-14}
			&GPT-3.5              &60.12   &56.69    &85.71   &68.24        &60.41   &56.16   &93.99   &70.26       &64.48   &59.93   &90.24   &71.01\\
			&GPT-4  & 73.79 & 69.18 & \textbf{88.57} & 77.68 & 73.18 & 66.5  & \textbf{93.57} & 77.74 & 74.23 & 66.56 & \textbf{98.12} & 79.32 \\
			\midrule
			
			\multicolumn{2}{c}{ }  & \multicolumn{4}{c}{400/5}  & \multicolumn{4}{c}{800/10} & \multicolumn{4}{c}{1600/20}  \\
			\midrule
			\multicolumn{2}{c}{ } & Acc     &Pre       &Rec        & F1            & Acc     &Pre       &Rec        & F1            & Acc     &Pre       &Rec        & F1  \\
			\multirow{9}*{Paper}&CNN       & 72.21 & 75.57 & 72.21 & 73.85 & 76.20 & 74.09 & 76.20 & 75.13 & 79.33 & 77.48 & 79.33 & 78.41 \\
			&BiLSTM    & 74.69 & 78.10 & 74.69 & 76.35 & 77.28 & 79.06 & 77.28 & 78.16 & 80.40 & 80.36 & 80.40 & 80.38 \\
			&GATED-CNN & 68.44 & 72.40 & 68.63 & 67.29 & 71.41 & 74.78 & 69.36 & 65.84 & 71.79 & 72.02 & 71.04 & 72.91 \\
			&FAST-TEXT & 67.93 & 70.25 & 66.03 & 65.53 & 68.45 & 72.88 & 68.96 & 67.85 & 71.10 & 72.15 & 72.23 & 70.39 \\
			&BERT      & 75.79 & 76.99 & 75.79 & 76.38 & 79.25 & 77.63 & 79.25 & 78.42 & 82.65 & 82.71 & 82.65 & 82.68\\
			&MFWCD-BERT           &\textbf{78.75}   &80.01   &\textbf{78.69}   &\textbf{78.90}         &\textbf{80.84}   &\textbf{80.83}   &80.79   &\textbf{81.03}       &\textbf{84.54}   &\textbf{85.02}  &84.11   &\textbf{84.51}\\
			&PT                   &72.07   &76.98   &72.24   &71.33         &75.61   &76.05   &75.61   &75.51       &80.41   &80.55  &80.41   &80.39\\
			&PEPL                 &70.14   &74.46   &70.14   &70.53         &73.62   &73.88   &73.62   &73.55       &78.28   &78.77  &78.28   &78.19\\
			\cdashline{2-14}
			&GPT-3.5              &68.33   &72.18   &62.70   &66.42         &68.32   &70.40   &63.88   &66.98       &70.94   &72.14   &67.21 &69.59\\
			&GPT-4  & 77.75 & \textbf{80.04} & 73.85 & 76.82 & 72.6  & 66.99 & \textbf{89.09} & 76.48 & 62.28 & 57.34 & \textbf{95.82} & 71.75\\
			\bottomrule
			
		\end{tabular}
	}
	%\label{tab:booktabs}
\end{table*}

\subsection{Experimental Results}

The main results of the experiments on clickbait detection in English and Chinese datasets are listed in Tables 3 and 4 respectively, which presents the detection results of ChatGPT (including GPT-3.5 and GPT-4) in the few-shot scenario, along with a comparison to other state-of-the-art methods. We can see that the performance of ChatGPT has not demonstrated the best results on all four metrics over the seven datasets. Compared to the deep neural networks and fine-tuning PLMs, ChatGPT has a lot of room for improvement in clickbait detection, especially the performance of GPT-3.5 on the Chinese dataset. In addition, with the increase of pre-trained information, the effectiveness of clickbait detection, sometimes, is worse. Therefore, we can observe that a small amount of pre-training information does not significantly impact the performance of ChatGPT in detecting clickbait. Moreover, we can observe that ChatGPT can achieve stable results on almost all evaluation metrics. The results confirmed that LLMs can be adapted to other languages, and ChatGPT is a monolithic model capable of supporting multiple languages, which makes it a comprehensive multilingual clickbait detection technique. It is worth noting that GPT-4 shows significant improvements over GPT-3.5 in areas such as comprehension and handling complex tasks, especially in scenarios that require a high level of text understanding, such as clickbait detection tasks.

Moreover, we have provided a detailed analysis of four news headlines where ChatGPT correctly classified them using the prompt method but made errors when detecting clickbait. For example, given the English not-clickbait headline "Tigers 'starved to death' to make \$500 aphrodisiac wine with their bones," ChatGPT misclassified it as clickbait. This is primarily because, first, the headline touches on sensitive topics such as animal abuse, illegal wildlife trade, and ethical controversies, which can easily elicit emotional responses from readers. Secondly, terms like "starved to death" and "aphrodisiac" are strong verbs and adjectives often used exaggeratedly or emotionally, and these words may match common vocabulary found in clickbait articles. Thirdly, in the absence of additional context, ChatGPT may be more inclined to categorize sentences with controversial or extreme characteristics as clickbait.For the Chinese not-clickbait headline "扬州通报“网传副市长与副局长生活作风问题”：建议对二人免职” (Yangzhou reports 'alleged misconduct in the personal conduct of deputy mayor and deputy director': Suggests dismissal of the two individuals.), ChatGPT misclassified it as clickbait. This is mainly because it is an attention-grabbing topic with relatively complex grammatical structures, including multiple parallel phrases and verb phrases. This complexity can lead the model to struggle in interpreting the sentence. It's also important to note that the phrase "网传" (spread on the internet) may have contributed to ChatGPT's classification as spreading false information.For the English clickbait headline "This is America's favorite fast food restaurant," ChatGPT misclassified it as not-clickbait. This is primarily because it is a topic related to American food culture and consumption habits, falling under the category of everyday life. Furthermore, the headline uses proper grammatical structure to pose a clear question, making it appear as a reasonable query rather than clickbait.For the Chinese clickbait headline "与你有关！一批重要国家标准，今天发布！" (Related to You! A batch of important national standards released today!), ChatGPT misclassified it as not-clickbait. This is mainly because national standards and regulations are a topic of practical significance, and the proper grammatical structure is used to present a statement. These factors may have led ChatGPT to mistakenly classify it as a legitimate news headline rather than clickbait.

In general, ChatGPT performs well in handling objective questions with clear, standardized answers. However, when it comes to addressing personalized questions, which often lack a fixed standard answer, it can be influenced by certain specific features. Moreover, conducting multiple tests on the same news may yield different results, leading to suboptimal performance in metrics like accuracy for clickbait detection.

\subsection{Ablation Study}

We compared the results between different prompts (as shown in Table 12) for clickbait detection tasks. As same as the main experiments, we select GPT-3.5 as a backbone for the prompts experiments. The results comparison of different Prompts are shown in Table 5.

Generally speaking, there is no significant performance gap between the first prompt (P1) and the second prompt (P2) in terms of effectiveness for clickbait detection. Specifically, the performance of P1 is significantly better than that of P2 in the dataset WeChat. In dataset Sina, the opposite is true.

\begin{table*}[htbp]
	\caption{Comparison of different prompts for GPT-3.5(\%). The bolder ones mean better.}
	\centering
	\resizebox{1.0\linewidth}{!}{
		\begin{tabular}{clcccc|cccc|cccc}
			\toprule
			Data & Method  & \multicolumn{12}{c}{Shot}  \\
			\cline{3-14}
			\multicolumn{2}{c}{ }  & \multicolumn{4}{c}{5}  & \multicolumn{4}{c}{10} & \multicolumn{4}{c}{20}  \\
			\midrule
			\multicolumn{2}{c}{ } & \multicolumn{12}{c}{Comparison of different prompts for GPT3.5 on English datasets} \\
			\cline{3-14}
			\multicolumn{2}{c}{ } & Acc     &Pre       &Rec        & F1            & Acc     &Pre       &Rec        & F1            & Acc     &Pre       &Rec        & F1  \\
			
			\multirow{2}*{DLC}&GPT3.5+P1&\textbf{80.75}   &\textbf{69.06}   &78.69   &\textbf{73.56}         &\textbf{79.33}   &\textbf{69.51}   &70.08   &69.79       &\textbf{82.07}   &\textbf{75.66}   &70.08 &72.76\\
			&GPT3.5+P2               &67.98   &52.20   &\textbf{82.79}   &64.03         &75.25   &59.33   &\textbf{88.02}   &\textbf{70.88}       &79.80   &66.45   &\textbf{82.64} &\textbf{73.67}\\
			\midrule
			\multirow{2}*{CND}&GPT3.5+P1&58.41   &\textbf{59.02}   &49.36   &53.76         &\textbf{59.44}   &\textbf{60.76}   &50.08   &54.91       &\textbf{60.90}   &\textbf{63.33}   &49.20 &55.38\\
			&GPT3.5+P2               &\textbf{60.00}   &56.83   &\textbf{68.36}   &\textbf{62.06}         &59.00   &59.28   &\textbf{61.00}   &\textbf{60.13}       &60.63   &58.78   &\textbf{66.88} &\textbf{62.57}\\
			\midrule
			\multirow{2}*{SCC}&GPT3.5+P1&\textbf{83.94}   &\textbf{88.63}   &78.19   &\textbf{83.08}         &\textbf{86.14}   &\textbf{92.41}   &78.85   &\textbf{85.09}       &\textbf{88.73}   &\textbf{95.54}   &81.26 &\textbf{87.82}\\
			&GPT3.5+P2               &81.49   &76.54   &\textbf{90.01}   &82.73         &84.39   &82.71   &\textbf{86.93}   &84.77       &87.52   &87.30   &\textbf{87.81} &87.55\\
			
			\bottomrule
			\multicolumn{2}{c}{ } & \multicolumn{12}{c}{Comparison of different prompts for GPT-3.5 on Chinese datasets} \\
			\cline{3-14}
			\multicolumn{2}{c}{ } & Acc     &Pre       &Rec        & F1            & Acc     &Pre       &Rec        & F1            & Acc     &Pre       &Rec        & F1  \\
			\multirow{2}*{Sina}&GPT3.5+P1&65.33   &68.80   &56.10   &61.80         &60.80   &60.69   &61.32   &61.00       &63.94   &64.60   &61.67 &63.10\\
			&GPT3.5+P2               &\textbf{70.73}   &\textbf{76.21}   &\textbf{60.28}   &\textbf{67.32}         &\textbf{69.16}   &\textbf{71.65}   &\textbf{63.41}   &\textbf{67.28}       &\textbf{75.30}   &\textbf{83.80}   &\textbf{62.85} &\textbf{0.7183}\\
			\midrule
			\multirow{2}*{Tencent}&GPT3.5+P1&\textbf{59.34}   &\textbf{57.91}   &\textbf{71.13}   &\textbf{63.84}         &52.88   &52.89   &52.62   &52.75       &58.51   &\textbf{57.17}   &67.80 &62.03\\
			&GPT3.5+P2               &58.64   &56.96   &70.68   &63.08         &\textbf{56.02}   &\textbf{54.71}   &\textbf{69.99}   &\textbf{61.38}       &\textbf{59.08}   &57.17   &\textbf{71.99} &\textbf{63.73}\\
			\midrule
			\multirow{2}*{WeChat}&GPT3.5+P1&\textbf{60.12}   &\textbf{56.69}   &\textbf{85.71}   &\textbf{68.24}         &\textbf{60.41}   &56.10   &\textbf{93.99}   &\textbf{70.26}       &58.57   &55.25   &\textbf{90.24} &68.54\\
			&GPT3.5+P2               &55.71   &53.99   &77.38   &63.60         &58.69   &\textbf{56.16}   &79.29   &65.75       &\textbf{64.48}   &\textbf{59.93}   &87.11 &\textbf{71.01}\\
			\midrule
			\multirow{2}*{Paper}&GPT3.5+P1&\textbf{68.33}   &70.60   &\textbf{62.70}   &\textbf{66.42}         &\textbf{68.32}   &\textbf{70.40}   &\textbf{63.88}   &\textbf{66.98}       &68.70   &70.84   &61.94 &66.09\\
			&GPT3.5+P2               &68.13   &\textbf{72.18}   &59.59   &65.28        &66.43   &69.54   &55.77   &61.9       &\textbf{70.94}   &\textbf{72.14}   &\textbf{67.21} &\textbf{69.59}\\
			\bottomrule
		\end{tabular}
	}
	
	%\label{tab:booktabs}
\end{table*}

\subsection{The Impact of Different LLMs}
Due to the rapid development and diverse characteristics of LLMs, we selected several different LLMs as the foundation for our experiments on clickbait detection. Specifically, we chose local LLMs including LLaMA3-8B, Yi-34B, and DeepSeek-32B, and compared their performance with commercial models such as GPT-3.5 and GPT-4. The results are presented in Tables 6 and 7.

We observe that GPT-4 consistently performs well across most datasets. Among the local LLMs, DeepSeek-32B demonstrates relatively stable performance and even surpasses GPT-3.5 in certain metrics. However, it is important to note that some LLMs tend to favor specific classes, resulting in a significant imbalance between precision and recall. As a result, LLMs still show considerable limitations in the task of clickbait detection.

% \usepackage{multirow}
\begin{table*}[htbp]
	\caption{The Accuracy, Precision, Recall, F1-scores results (\%) on English datasets with different LLMs(\%). The bolder ones mean better.}
	\centering
	\resizebox{1.0\linewidth}{!}{
		\begin{tabular}{lllll|llll|llll}
			\toprule
			& \multicolumn{4}{c}{DLC}                                        & \multicolumn{4}{c}{CND}                                          & \multicolumn{4}{c}{SCC}       \\
			\cline{2-13}
			& Acc            & Pre         & Rec            & F1             & Acc            & Pre            & Rec            & F1            & Acc   & Pre   & Rec   & F1    \\
			LLaMA3   & 77.68          & 66.41          & 69.67          & 68.00          & 59.32          & 61.52          & 49.77          & 55.02          & 82.95          & 88.78          & 75.43          & 81.56          \\
			Yi       & 76.71          & 82.91          & 39.75          & 53.74          & 58.19          & 58.81          & 54.66          & 56.66          & 79.10          & 84.12          & 70.58          & 76.92          \\
			DeepSeek & 72.11          & 57.80          & 66.80          & 61.98          & 62.85          & 62.53          & 64.13          & 63.32          & 84.13          & 90.22          & 76.57          & 82.83          \\
			GPT-3.5  & 80.75          & 69.06          & \textbf{82.79} & 73.56          & 60.00          & 59.02          & \textbf{68.36} & 62.06          & 83.94          & 88.63          & 90.01          & 83.08          \\
			GPT-4    & \textbf{90.54} & \textbf{94.00} & 77.05          & \textbf{84.68} & \textbf{68.61} & \textbf{69.47} & 66.97          & \textbf{68.20} & \textbf{94.62} & \textbf{98.83} & \textbf{90.30} & \textbf{94.37}\\
			\bottomrule
	\end{tabular}}
\end{table*}

\begin{table*}[htbp]
	\caption{The Accuracy, Precision, Recall, F1-scores results (\%) on Chinese datasets with different LLMs(\%). The bolder ones mean better.}
	\centering
	\resizebox{1.1\linewidth}{!}{
		\begin{tabular}{lllll|llll|llll|llll}
			\toprule
			& \multicolumn{4}{c}{Sina}                                          & \multicolumn{4}{c}{Tencent}                                       & \multicolumn{4}{c}{WeChat}                                        & \multicolumn{4}{c}{Paper}                                         \\
			\cline{2-17}
			& Acc            & Pre            & Rec            & F1             & Acc            & Pre            & Rec            & F1             & Acc            & Pre            & Rec            & F1             & Acc            & Pre            & Rec            & F1             \\
			LLaMA3   & 73.52          & 69.12          & \textbf{85.02} & 76.25          & 63.87          & 60.04          & \textbf{82.98} & 69.67          & 54.64          & 52.91          & 84.52          & 65.08          & 70.63          & 64.42          & \textbf{92.16} & 75.83          \\
			Yi       & 68.89          & 71.07          & 69.97          & 70.55          & 72.54          & 68.87          & 73.46          & 71.08          & 63.68          & 61.02          & 75.71          & 67.52          & 76.45          & 76.54          & 76.30          & 76.30          \\
			DeepSeek & \textbf{82.82} & \textbf{93.67} & 76.83          & \textbf{84.42} & \textbf{78.45} & \textbf{77.35} & 80.47          & \textbf{79.03} & \textbf{74.66} & \textbf{71.01} & 83.33          & 76.66          & \textbf{79.39} & \textbf{88.32} & 74.70          & \textbf{80.89} \\
			GPT-3.5  & 70.73          & 76.21          & 60.28          & 67.32          & 59.34          & 57.91          & 71.13          & 63.84          & 60.12          & 56.69          & 85.71          & 68.24          & 68.33          & 72.18          & 62.70          & 66.42          \\
			GPT-4    & 73.87          & 93.08          & 75.57          & 76.96          & 65.97          & 67.76          & 75.94          & 71.58          & 73.79          & 69.18          & \textbf{88.57} & \textbf{77.68} & 77.75          & 80.04          & 73.85          & 76.82         
			\\
			\bottomrule
	\end{tabular}}
\end{table*}

\section{Conclusion and Future Work}

In this paper, we present a study of the performance of LLMs (ChatGPT with GPT-3.5 and GPT-4) for clickbait detection. During the benchmark experiments on both English and Chinese datasets, LLMs did not perform as well as the current state-of-the-art deep neural networks and fine-tuning PLMs methods in the realm of multilingual clickbait detection. In our subsequent efforts, we will try to design more effective methods with the help of LLMs that can significantly and consistently outperform SOTA clickbait detection methods.

\bibliographystyle{named}
\bibliography{ijcai23}

\begin{thebibliography}{}

\bibitem[\protect\citeauthoryear{Agrawal}{2016}]{agrawal2016clickbait}
Amol Agrawal.
\newblock Clickbait detection using deep learning.
\newblock In {\em 2016 2nd international conference on next generation
  computing technologies (NGCT)}, pages 268--272. IEEE, 2016.

\bibitem[\protect\citeauthoryear{Anand \bgroup \em et al.\egroup
  }{2017}]{anand2017we}
Ankesh Anand, Tanmoy Chakraborty, and Noseong Park.
\newblock We used neural networks to detect clickbaits: You won't believe what
  happened next!
\newblock In {\em Proceedings of the European Conference on Information
  Retrieval}, pages 541--547, 2017.

\bibitem[\protect\citeauthoryear{Biyani \bgroup \em et al.\egroup
  }{2016}]{biyani20168}
Prakhar Biyani, Kostas Tsioutsiouliklis, and John Blackmer.
\newblock "8 amazing secrets for getting more clicks": detecting clickbaits in
  news streams using article informality.
\newblock In {\em Proceedings of the AAAI Conference on Artificial Intelligence
  (AAAI)}, pages 94--100, 2016.

\bibitem[\protect\citeauthoryear{Blom and Hansen}{2015}]{blom2015click}
Jonas~Nygaard Blom and Kenneth~Reinecke Hansen.
\newblock Click bait: Forward-reference as lure in online news headlines.
\newblock {\em Journal of Pragmatics}, 76:87--100, 2015.

\bibitem[\protect\citeauthoryear{Brown \bgroup \em et al.\egroup
  }{2020}]{brown2020language}
Tom~B Brown, Benjamin Mann, Nick Ryder, Melanie Subbiah, Jared Kaplan, Prafulla
  Dhariwal, Arvind Neelakantan, Pranav Shyam, Girish Sastry, Amanda Askell,
  et~al.
\newblock Language models are few-shot learners.
\newblock In {\em Neural Information Processing Systems}, pages 1877--1901,
  2020.

\bibitem[\protect\citeauthoryear{Chakraborty \bgroup \em et al.\egroup
  }{2016}]{chakraborty2016stop}
Abhijnan Chakraborty, Bhargavi Paranjape, Sourya Kakarla, and Niloy Ganguly.
\newblock Stop clickbait: Detecting and preventing clickbaits in online news
  media.
\newblock In {\em Proceedings of the IEEE/ACM International Conference on
  Advances in Social Networks Analysis and Mining (ASONAM)}, pages 9--16, 2016.

\bibitem[\protect\citeauthoryear{Chen \bgroup \em et al.\egroup
  }{2015}]{chen2015misleading}
Yimin Chen, Niall~J Conroy, and Victoria~L Rubin.
\newblock Misleading online content: recognizing clickbait as "false news".
\newblock In {\em Proceedings of the ACM on workshop on Multimodal Deception
  Detection}, pages 15--19, 2015.

\bibitem[\protect\citeauthoryear{Chowdhery \bgroup \em et al.\egroup
  }{2022}]{chowdhery2022palm}
Aakanksha Chowdhery, Sharan Narang, Jacob Devlin, Maarten Bosma, Gaurav Mishra,
  Adam Roberts, Paul Barham, Hyung~Won Chung, Charles Sutton, Sebastian
  Gehrmann, et~al.
\newblock Palm: Scaling language modeling with pathways.
\newblock {\em arXiv preprint arXiv:2204.02311}, 2022.

\bibitem[\protect\citeauthoryear{Devlin \bgroup \em et al.\egroup
  }{2018}]{devlin2018bert}
Jacob Devlin, Ming-Wei Chang, Kenton Lee, and Kristina Toutanova.
\newblock Bert: Pre-training of deep bidirectional transformers for language
  understanding.
\newblock {\em arXiv preprint arXiv:1810.04805}, 2018.

\bibitem[\protect\citeauthoryear{Indurthi \bgroup \em et al.\egroup
  }{2020}]{indurthi2020predicting}
Vijayasaradhi Indurthi, Bakhtiyar Syed, Manish Gupta, and Vasudeva Varma.
\newblock Predicting clickbait strength in online social media.
\newblock In {\em Proceedings of the International Conference on Computational
  Linguistics}, pages 4835--4846, 2020.

\bibitem[\protect\citeauthoryear{Joulin \bgroup \em et al.\egroup
  }{2016}]{joulin2016bag}
Armand Joulin, Edouard Grave, Piotr Bojanowski, and Tomas Mikolov.
\newblock Bag of tricks for efficient text classification.
\newblock {\em arXiv preprint arXiv:1607.01759}, 2016.

\bibitem[\protect\citeauthoryear{Kumar \bgroup \em et al.\egroup
  }{2018}]{kumar2018identifying}
Vaibhav Kumar, Dhruv Khattar, Siddhartha Gairola, Yash Kumar~Lal, and Vasudeva
  Varma.
\newblock Identifying clickbait: A multi-strategy approach using neural
  networks.
\newblock In {\em The 41st International ACM SIGIR Conference on Research \&
  Development in Information Retrieval}, pages 1225--1228, 2018.

\bibitem[\protect\citeauthoryear{Liu \bgroup \em et al.\egroup
  }{2019}]{liu2019roberta}
Yinhan Liu, Myle Ott, Naman Goyal, Jingfei Du, Mandar Joshi, Danqi Chen, Omer
  Levy, Mike Lewis, Luke Zettlemoyer, and Veselin Stoyanov.
\newblock Roberta: A robustly optimized bert pretraining approach.
\newblock {\em arXiv preprint arXiv:1907.11692}, 2019.

\bibitem[\protect\citeauthoryear{Liu \bgroup \em et al.\egroup
  }{2022}]{liu2022clickbait}
Tong Liu, Ke~Yu, Lu~Wang, Xuanyu Zhang, Hao Zhou, and Xiaofei Wu.
\newblock Clickbait detection on wechat: A deep model integrating semantic and
  syntactic information.
\newblock {\em Knowledge-Based Systems}, 245:108605, 2022.

\bibitem[\protect\citeauthoryear{Liu \bgroup \em et al.\egroup
  }{2023}]{liu2023pre}
Pengfei Liu, Weizhe Yuan, Jinlan Fu, Zhengbao Jiang, Hiroaki Hayashi, and
  Graham Neubig.
\newblock Pre-train, prompt, and predict: A systematic survey of prompting
  methods in natural language processing.
\newblock {\em ACM Computing Surveys}, 55(9):1--35, 2023.

\bibitem[\protect\citeauthoryear{L{\'o}pez-S{\'a}nchez \bgroup \em et
  al.\egroup }{2018}]{lopez2018hybridizing}
Daniel L{\'o}pez-S{\'a}nchez, Jorge~Revuelta Herrero, Ang{\'e}lica~Gonz{\'a}lez
  Arrieta, and Juan~M Corchado.
\newblock Hybridizing metric learning and case-based reasoning for adaptable
  clickbait detection.
\newblock {\em Applied Intelligence}, 48(9):2967--2982, 2018.

\bibitem[\protect\citeauthoryear{Mishra \bgroup \em et al.\egroup
  }{2020}]{mishra2020musem}
Rahul Mishra, Piyush Yadav, Remi Calizzano, and Markus Leippold.
\newblock Musem: Detecting incongruent news headlines using mutual attentive
  semantic matching.
\newblock In {\em Proceedings of the IEEE International Conference on Machine
  Learning and Applications (ICMLA)}, pages 709--716, 2020.

\bibitem[\protect\citeauthoryear{Raffel \bgroup \em et al.\egroup
  }{2020}]{raffel2020exploring}
Colin Raffel, Noam Shazeer, Adam Roberts, Katherine Lee, Sharan Narang, Michael
  Matena, Yanqi Zhou, Wei Li, and Peter~J Liu.
\newblock Exploring the limits of transfer learning with a unified text-to-text
  transformer.
\newblock {\em The Journal of Machine Learning Research}, 21(1):5485--5551,
  2020.

\bibitem[\protect\citeauthoryear{Rony \bgroup \em et al.\egroup
  }{2017}]{rony2017diving}
Md~Main~Uddin Rony, Naeemul Hassan, and Mohammad Yousuf.
\newblock Diving deep into clickbaits: Who use them to what extents in which
  topics with what effects?
\newblock In {\em Proceedings of the IEEE/ACM international conference on
  advances in social networks analysis and mining 2017}, pages 232--239, 2017.

\bibitem[\protect\citeauthoryear{Thoppilan \bgroup \em et al.\egroup
  }{2022}]{thoppilan2022lamda}
Romal Thoppilan, Daniel De~Freitas, Jamie Hall, Noam Shazeer, Apoorv
  Kulshreshtha, Heng-Tze Cheng, Alicia Jin, Taylor Bos, Leslie Baker, Yu~Du,
  et~al.
\newblock Lamda: Language models for dialog applications.
\newblock {\em arXiv preprint arXiv:2201.08239}, 2022.

\bibitem[\protect\citeauthoryear{Wei and Wan}{2017}]{wei2017learning}
Wei Wei and Xiaojun Wan.
\newblock Learning to identify ambiguous and misleading news headlines.
\newblock pages 4172--4178, 2017.

\bibitem[\protect\citeauthoryear{Wu \bgroup \em et al.\egroup
  }{2023}]{wu2023detecting}
Yin Wu, Mingpei Cao, Yueze Zhang, and Yong Jiang.
\newblock Detecting clickbait in chinese social media by prompt learning.
\newblock In {\em 2023 26th International Conference on Computer Supported
  Cooperative Work in Design (CSCWD)}, pages 369--374. IEEE, 2023.

\bibitem[\protect\citeauthoryear{Yang \bgroup \em et al.\egroup
  }{2019}]{yang2019fake}
Kai-Chou Yang, Timothy Niven, and Hung-Yu Kao.
\newblock Fake news detection as natural language inference.
\newblock {\em arXiv preprint arXiv:1907.07347}, 2019.

\bibitem[\protect\citeauthoryear{Yi \bgroup \em et al.\egroup
  }{2022}]{yi2022clickbait}
Xiaoyuan Yi, Jiarui Zhang, Wenhao Li, Xiting Wang, and Xing Xie.
\newblock Clickbait detection via contrastive variational modelling of text and
  label.
\newblock In {\em Proceedings of the Thirty-First International Joint
  Conference on Artificial Intelligence (IJCAI)}, pages 4475--4481, 2022.

\bibitem[\protect\citeauthoryear{Yoon \bgroup \em et al.\egroup
  }{2019}]{yoon2019detecting}
Seunghyun Yoon, Kunwoo Park, Joongbo Shin, Hongjun Lim, Seungpil Won, Meeyoung
  Cha, and Kyomin Jung.
\newblock Detecting incongruity between news headline and body text via a deep
  hierarchical encoder.
\newblock In {\em Proceedings of the AAAI Conference on Artificial Intelligence
  (AAAI)}, volume~33, pages 791--800, 2019.

\bibitem[\protect\citeauthoryear{Zheng \bgroup \em et al.\egroup
  }{2021}]{zheng2021deep}
Jiaming Zheng, Ke~Yu, and Xiaofei Wu.
\newblock A deep model based on lure and similarity for adaptive clickbait
  detection.
\newblock {\em Knowledge-Based Systems}, 214:106714, 2021.

\end{thebibliography}

\end{CJK}
\end{document}